\documentclass[runningheads]{llncs}
\usepackage{graphicx}
\usepackage{cite}
\usepackage{textcomp}
\usepackage{xcolor}
\usepackage{algorithm}
\usepackage[noend]{algpseudocode}
\algrenewcommand{\algorithmiccomment}[1]{\hskip1em$\#$ #1}
\usepackage{url}
\usepackage{multirow}
\usepackage{xcolor}

\begin{document}

\title{\large Automatically Balancing Model Accuracy and Complexity using Solution and Fitness Evolution (SAFE)\thanks{This work was supported by National Institutes of Health (USA) grants LM010098 and AG066833.}}

\titlerunning{Automatically Balancing Model Accuracy and Complexity}

\author{Moshe Sipper\inst{1,2} \and
Jason H. Moore\inst{1} \and
Ryan J. Urbanowicz\inst{1}}
\authorrunning{M. Sipper et al.}


\institute{\scriptsize Institute for Biomedical Informatics, University of Pennsylvania, Philadelphia, PA 19104, USA \and
Department of Computer Science, Ben-Gurion University, Beer Sheva 84105, Israel \\
\email{sipper@gmail.com},
\url{https://epistasis.org/} }

\maketitle

\begin{abstract}
When seeking a predictive model in biomedical data, one often has more than a single objective in mind, e.g., attaining both high accuracy and low complexity (to promote interpretability).
We  investigate herein  whether  multiple  objectives  can  be  dynamically tuned by our recently proposed coevolutionary algorithm, SAFE (Solution And Fitness Evolution).
We find that SAFE is able to automatically tune accuracy and complexity with no performance loss, as compared with a standard evolutionary algorithm, over complex simulated genetics datasets produced by the GAMETES tool.

\end{abstract}

\section{Introduction}
\label{sec:intro}
One of the bread-and-butter tasks in precision medicine is that of finding a predictive model for a given dataset. Commonly, a dataset will include various features---genetic, physiological, historical, lifestyle---pertaining to a particular outcome (condition, syndrome, disease). The modeler's job is then to parlay this dataset into a useful model.

A useful model will predict unknown outcomes as accurately as possible, e.g., exhibiting as few false negatives/positives when given a putative case of cancer, with the objective of deciding whether it is benign or malignant. Often, though, objectives other than sheer accuracy will be of interest. For example, it might be beneficial to seek a model that is not only accurate but also simple---and thus more amenable to interpretation. Identifying the right balance of accuracy and complexity is also important in order to avoid overfitting, particularly when the optimal expected accuracy is unknown (i.e., how much true signal actually exists in the dataset).

If more than one objective is to be sought the field of multi- and many-objective evolutionary optimization has much to offer\cite{zhou2011multiobjective,Bezerra2018,deb2002fast} (and indeed we have applied the SAFE algorithm studied herein to multiobjective optimization\cite{Sipper2019multiobj}). However, these approaches come with various costs, assumptions, and restrictions (the enumeration of which is beyond the scope of this paper).

Herein we investigate whether multiple objectives can be dynamically tuned, while remaining within the realm of single-objective optimization. To wit, we present an evolutionary algorithm that ultimately seeks a single model rather than a plethora of models along a Pareto front (as done with multiobjective optimization). 

In order to properly test this concept, this investigation relies on the generation of a diverse simulation study that affords us the opportunity to control the complexity and signal over a broad set of scenarios. To this end, we have opted to utilize a tool called GAMETES \cite{urbanowicz2012gametes} that allows us to rapidly generate complex biallelic single nucleotide polymorphism (SNP) disease models and associated simulated datasets. 

Our aim is to find models that are both accurate and simple for datasets produced by GAMETES, which span a wide range of constraints. Toward this end we employ two evolutionary algorithms (EA). The first is a standard EA that combines model accuracy and model complexity into a single fitness function through fixed weighting. The second EA is our recently developed SAFE algorithm,\cite{Sipper2019robot,Sipper2019multiobj} which automatically tunes both accuracy and complexity by harnessing the potential of \textit{coevolution}. This is in-line with the concept of data/problem driven model optimization. 

The study presented herein is still work-in-progress.
The main question we ask is:\textit{Within a scenario of model discovery can we forgo the need to a-priori perform a balancing act between multiple objectives?}
To wit, our main aim is to show that one can forgo weighting components of the objective function, letting these be balanced automatically---with no loss in performance. We believe this avenue of investigation to be potentially valuable to downstream large-scale biomedical data analysis, where it will become less computationally feasible to perform predictive modeling over an array of pre-defined accuracy/complexity trade-off weights.  

Section~\ref{sec:gametes} presents the GAMETES dataset generation tool. Section~\ref{sec:safe} presents the SAFE algorithm. Section~\ref{sec:results} presents our experimental setup and results, followed by concluding remarks and future directions in Section~\ref{sec:conc}.

\section{Genetic Architecture Model Emulator for Testing and Evaluating Software (GAMETES)} 
\label{sec:gametes}

The GAMETES software was designed to rapidly and `randomly' generate $n$-locus biallelic SNP models of disease that represent pure, strict examples of statistical epistasis \cite{urbanowicz2012gametes}. In other words, individual SNPs in these models have no independent main effects, nor any subset of `nested' SNPs in the overall $n$-locus model have an association with outcome (i.e., no subset of 2 SNPs within a 3-locus model are predictive). The models generated by GAMETES must also conform to the contraints specified by the user, including; heritability, individual SNP minor allele frequencies, population prevalence, and the order of interaction---$n$. 

GAMETES was extended with the ability to estimate model difficulty directly from the penetrance function it produced as the model \cite{urbanowicz2012predicting}. This allows researchers to rapidly generate a population of `random' models that each fit the standard model constraints defined by the user. We can then select models at the extremes of this distribution to capture the detection difficulty beyond the standard specified model constraints, which the authors referred to as `model architecture'. In the present study models with a maximally difficult architecture are designated as `EDM-1', while those with a maximally easy architecture are designated as `EDM-2'. Once models have been generated, GAMETES also includes a simple dataset simulation strategy to embed the specific simulated epistatic model in a dataset among other randomly generated non-predictive SNPs. 

Similar to previous applications of GAMETES designed to test novel machine learning methodologies in complex biomedical data tasks \cite{urbanowicz2015exstracs}, in the present study we applied GAMETES to generate 2-locus and 3-locus datasets, over a range of heritabilities (i.e., 0.1, 0.2, 0.3, 0.4), with minor allele frequencies fixed at 0.2, and model prevalence values allowed to vary during `random' model generation. For simplicity, in the present study we generated all datasets to have a sample size of 800, and a total of 100 features. 

Due to the psuedo-random model generation strategy utilized by GAMETES, it is nearly impossible to automatically generate models approaching a heritability of 1 in a reasonable amount of time. Therefore, we utilized GAMETES's `custom' model generation function within the software's graphical user interface to generate a 2-way and 3-way biallelic pure, strict, epistatic model with a heritability of 1, referred to in related work as the XOR model. 

The GAMETES version used in this study is available at \url{https://github.com/UrbsLab/GAMETES}. 

\section{Solution and Fitness Evolution (SAFE): Previous and Current Work} 
\label{sec:safe}
Coevolutionary algorithms simultaneously evolve two or more populations with coupled fitness \cite{Pena:2001}. Strongly related to the concept of symbiosis, coevolution can be mutualistic (cooperative), parasitic (competitive), or commensalistic:\footnote{\url{https://en.wikipedia.org/wiki/Symbiosis}} 
1) In cooperative coevolution, different species exist in a relationship in which each individual (fitness) benefits from the activity of the other; 2) in competitive coevolution, an organism of one species competes with an organism of a different species; and 3) with commensalism, members of one species gain benefits while those of the other species neither benefit nor are harmed.
The idea of coevolution originates (at least) with Darwin---who spoke of ``coadaptations of organic beings to each other'' \cite{Darwin:1859}.  


SAFE is a coevolutionary algorithm that maintains two coevolving populations: a population of candidate solutions and a population of candidate objective functions.
To date, we applied SAFE within two domains: evolving robot controllers to solve mazes \cite{Sipper2019robot} and multiobjective optimization \cite{Sipper2019multiobj}.

\begin{figure}
\centering
\begin{tabular}{c@{\hskip 20px}c}
\includegraphics[width=0.3\textwidth]{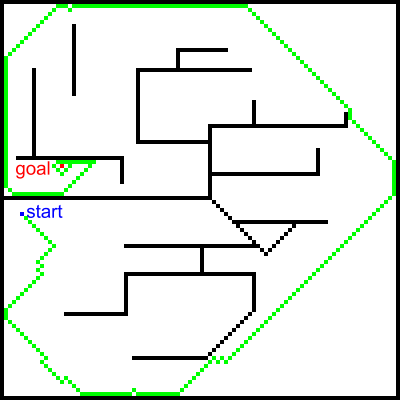} &
\includegraphics[width=0.3\textwidth]{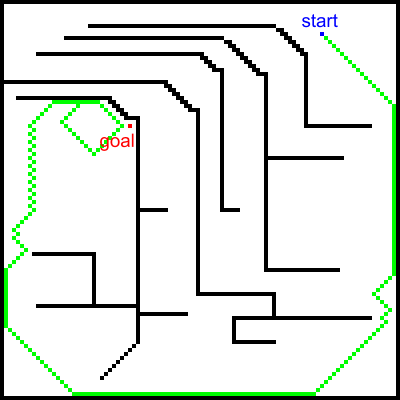} \\
\texttt{maze1} & \texttt{maze2}
\end{tabular}
\caption{Evolving robot controllers. Sample solutions to the maze problems, evolved by SAFE.} \label{fig:maze-solutions}
\end{figure}

Applying SAFE within the robotic domain,  an individual in the solutions population was a list of 16 real values, representing the robot's control vector (``brain''). The controller determined the robot's behavior when wandering a given maze, with its phenotype taken to be the final position, or endpoint. The endpoint was used to compute standard distance-to-goal fitness and to compute phenotypic novelty: compare the endpoint to all endpoints of current-generation robots \textit{and} to all endpoints in an archive of past individuals whose behaviors were highly novel when they emerged. The final novelty score was then the average of the 15 nearest neighbors. An individual in the objective-functions population was a list of 2 real values $[a,b]$, balancing `distance to goal' and novelty. 
SAFE performed far better than random search and a standard fitness-based evolutionary algorithm, and compared favorably with novelty search. Figure~\ref{fig:maze-solutions} shows sample solutions found by SAFE. For full details see \cite{Sipper2019robot}.

The second domain we applied SAFE to was multiobjective optmization \cite{Sipper2019multiobj}. With a multiobjective optimization problem there is usually no single-best solution, but rather the goal is to identify a set of `non-dominated' solutions that represent optimal tradeoffs between multiple objectives---the \textit{Pareto front}. Usually, a representative subset will suffice.
Specifically, we applied SAFE to the solution of 
the classical ZDT problems, which
epitomize the basic setup of multiobjective optimization
\cite{zitzler2000comparison,huband2006review}.
For example, ZDT1 is defined as:

\[ f_1(\textbf{x}) = x_1 \, ,\]

\[ g(\textbf{x}) = 1 + 9/(k-1)\sum_{i=2}^{k} x_i \, ,\]

\[ f_2(\textbf{x}) = 1-\sqrt{f_1/g} \, .\]
The two objectives are to minimize both $f_1(\textbf{x})$ and $f_2(\textbf{x})$.
The dimensionality of the problem is $k=30$, i.e.,
solution vector 
$\textbf{x} = x_1,\ldots,x_{30}$, $x_i \in [0,1]$. 
The utility of this suite is that the ground-truth optimal Pareto front can be computed and used to determine and compare multiobjective algorithm performance. 
SAFE maintained two coevolving populations. An individual in the solutions population was a list of 30 real values.
An individual in the objective-functions population was a list of 2 real values $[a,b]$, defining a candidate set of weights, balancing the two objectives of the ZDT functions:
$a$ determined $f_1$'s weighting and $b$ determined $f_2$'s weighting. 

\begin{figure}
\centering
\begin{tabular}{cc}
\includegraphics[height=0.35\textwidth]{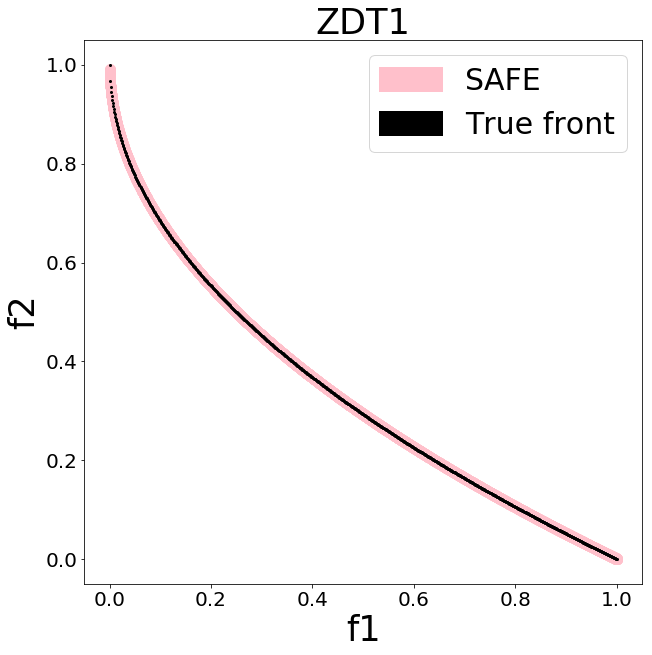} &
\includegraphics[height=0.35\textwidth]{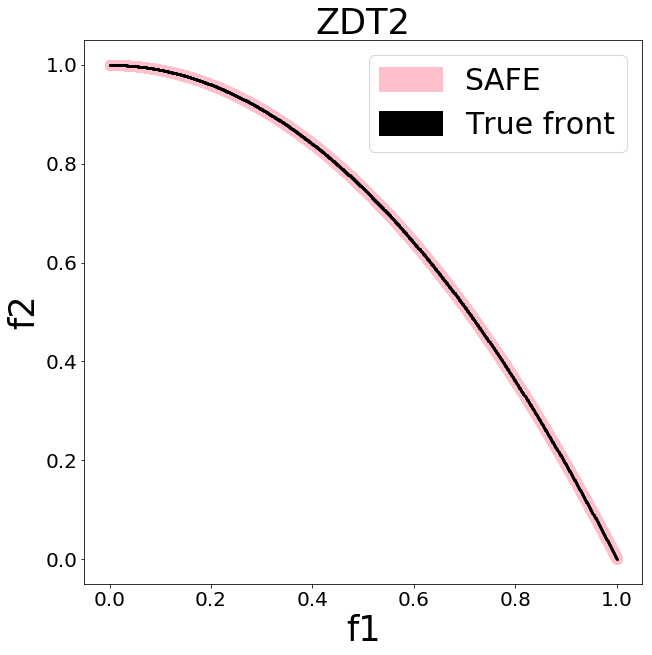} \\
\includegraphics[height=0.35\textwidth]{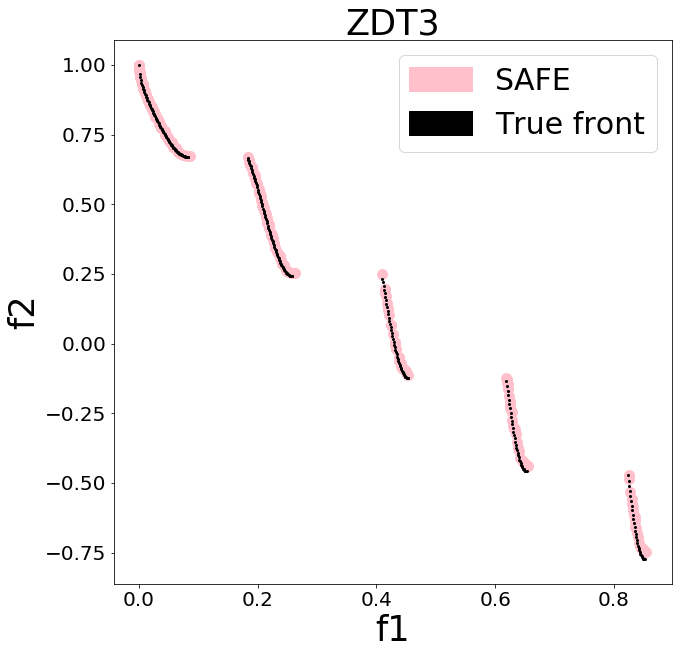} &
\includegraphics[height=0.35\textwidth]{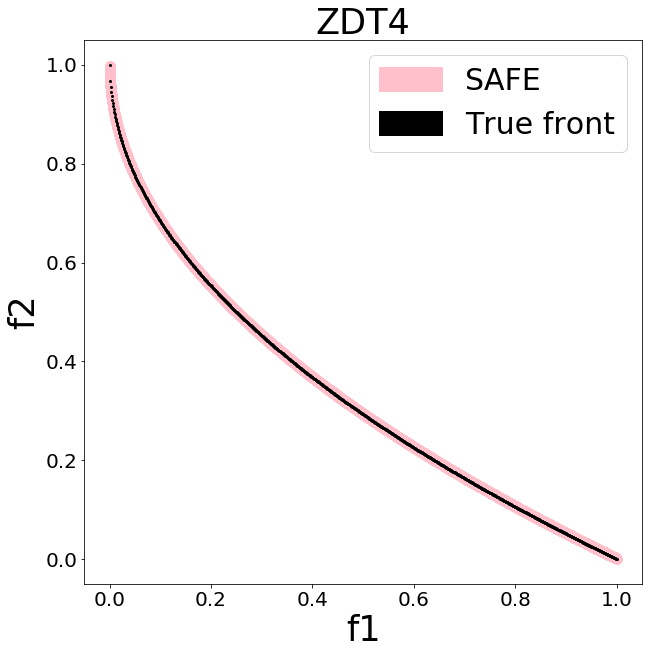} \\
\end{tabular}
\caption{Multiobjective optimization. Four sample solutions (Pareto fronts) produced by SAFE, compared with the true fronts.} \label{fig:fronts}
\end{figure}

We tested SAFE on four ZDT problems---ZDT1, ZDT2, ZDT3, ZDT4---recording the evolving Pareto front as evolution progressed. We compared our results with two recent studies
by Cheng et al. \cite{cheng2017novel} and by Han et al. \cite{han2018improved}, finding that SAFE was able to perform convincingly better on 3 of the 4 problems. Figure~\ref{fig:fronts} shows four Pareto fronts produced by SAFE, compared with the optimal ground truth Pareto fronts as defined by the ZDT functions. For full details see \cite{Sipper2019multiobj}.

We now turn to our current work. The evolution of each population is identical to a standard, single-population evolutionary algorithm---except where fitness computation is concerned. Below we describe the following components of the system: (1) populations, (2) initialization, (3) selection, (4) elitism, (5) crossover, (6) mutation, (7) fitness.

\textbf{Populations.}
SAFE maintains two coevolving populations. 
Whereas previously we employed linear genomes, herein we expanded SAFE to accommodate genetic programming (GP) computational trees. 

\begin{figure}
\centering
\includegraphics[width=0.85\textwidth]{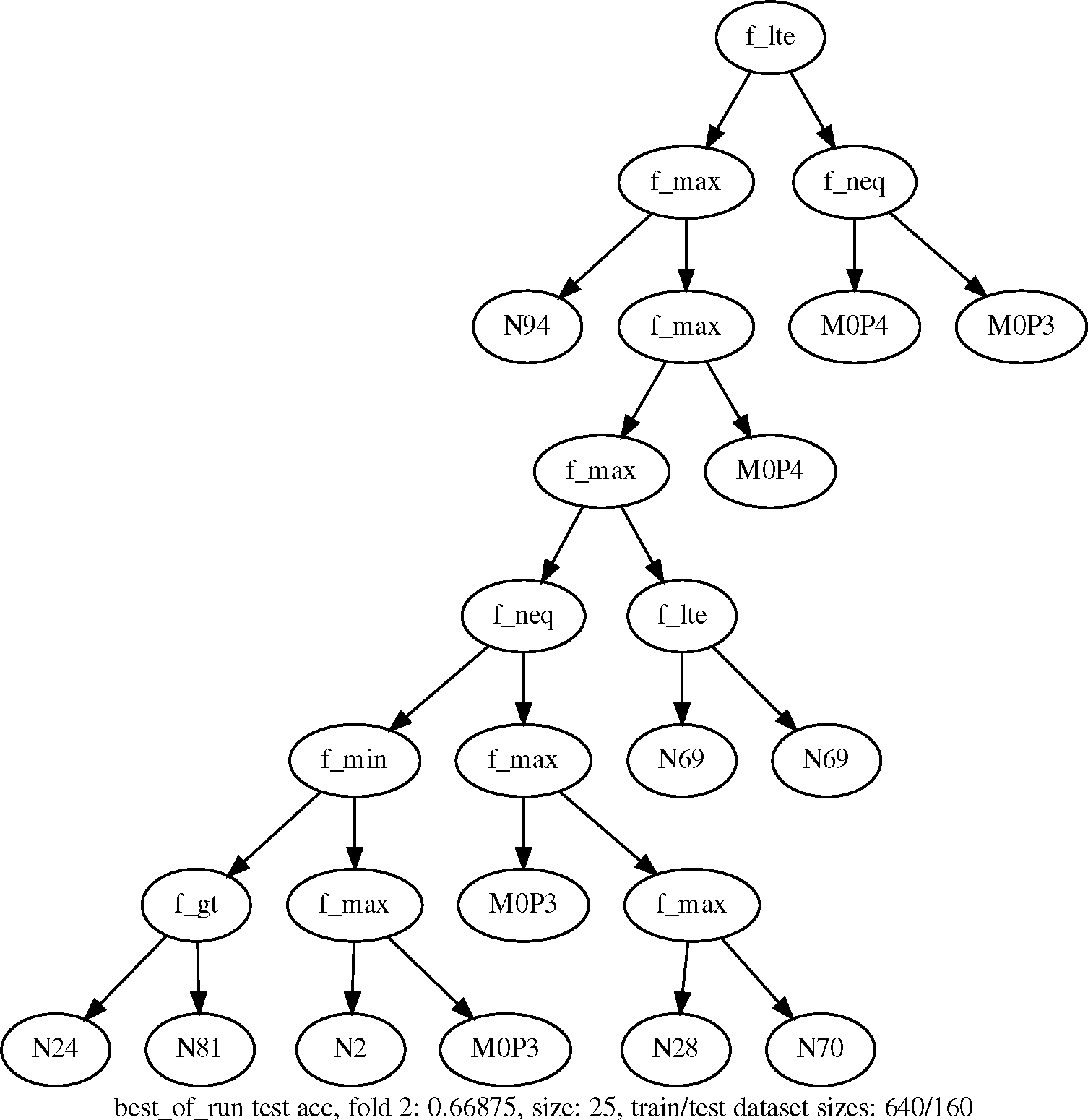} 
\caption{Sample evolved model tree.} 
\label{fig:model}
\end{figure}

An individual in the \emph{solutions population} is a model
tree composed of functions (internal nodes) and terminals (leaf nodes), which are specified in the function and terminal sets, respectively. 
The function set consists of arithmetic and relational operators
(Table~\ref{tab:gp}). The terminal set depends on the particular GAMETES dataset used, e.g., for the sample model tree shown in Figure~\ref{fig:model} this set includes 100 terminals: \texttt{\{N0, \ldots, N97, M0P3, M0P4\}}, with \texttt{M0P3} and \texttt{M0P4} being the predictive features.

\begin{table}
\centering
\caption{Function set.
NB: \texttt{f\_div} differs from the ubiquitous protected-division operator: \textit{x/y if y!=0 else c}; we used instead the better operator suggested by  \cite{ni2012use}.}
\small
{\begin{tabular}{@{}r@{\hskip 10px}l@{}}
Function              & Returns \\ \hline
\texttt{f\_add(x,y)}  & $x+y$ \\
\texttt{f\_sub(x,y)}  & $x-y$ \\
\texttt{f\_mul(x,y)}  & $x*y$ \\
\texttt{f\_div(x,y)}  & $x/\sqrt{1+y*y}$ \\
\texttt{f\_mean(x,y)} & $(x+y)/2$ \\
\texttt{f\_eq(x,y)}   & \textit{1 if x = y else -1} \\
\texttt{f\_neq(x,y)}  & \textit{1 if x $\neq$ y else -1} \\
\texttt{f\_lt(x,y)}   & \textit{1 if x $<$ y  else -1} \\
\texttt{f\_lte(x,y)}  & \textit{1 if x $\leq$ y else -1} \\
\texttt{f\_gt(x,y)}   & \textit{1 if x $>$ y  else -1} \\
\texttt{f\_gte(x,y)}  & \textit{1 if x $\geq$ y else -1} \\
\texttt{f\_min(x,y)}  & $\mathit{min}(x,y)$ \\
\texttt{f\_max(x,y)}  & $\mathit{max}(x,y)$ \\ 
\end{tabular}}\label{tab:gp}
\normalsize
\end{table}

An individual in the \emph{objective-functions population} is a list of 2 real values $[a,b]$, each in the range $[0,1]$, whose usage is described under \emph{fitness} below. Population sizes and other parameters are given in Table~\ref{tab:params} (Sipper et al. \cite{Sipper2018par} provide interesting insights into parameters in evolutionary computation).

\begin{table}
\centering
\caption{Evolutionary parameters (s: solutions, o: objective functions).}
\resizebox{0.4\textwidth}{!}{
{\begin{tabular}{@{}r@{\hskip 10px}l@{}}
Parameter          & Value           \\ \hline
No. replicate runs & 30              \\
No. generations    & 2000            \\
Population size    & 100 (s), 30 (o) \\
Selection          & Lexicase        \\
Crossover          & Single-point    \\
Crossover rate     & 0.8             \\
Mutation rate      & 0.4             \\
Elitism            & 2               \\
Generation gap     & 5
\\ 
\end{tabular}}\label{tab:params}
}
\end{table}

\textbf{Initialization.}
For every evolutionary run: 
the trees in the solutions population are initialized using the ramped-half-and-half method\cite{koza92}; the size-2 tuples in the objective-functions  population are initialized to random values in the appropriate range.

\textbf{Selection.} 
Whereas previously SAFE used tournament selection,
herein we implemented lexicase selection,\cite{metevier2019lexicase,spector2012assessment} which proved somewhat better. 
Lexicase selection selects individuals by filtering a pool of individuals which, before filtering, typically contains the entire population. The filtering is accomplished
in steps, each of which filters according to performance on a single test case.

\textbf{Elitism.} The 2 individuals with the highest fitness in a generation are copied (``cloned'') into the next generation unchanged.

\textbf{Crossover.} 
Subtree crossover for the solutions population, 
standard single-point crossover for the objective-functions population.

\textbf{Mutation.} 
Subtree mutation for the solutions population.
For the objective-functions population mutation is done with probability 0.4 (per individual in the population) by selecting a random gene of the 2 and replacing it with a new random value in the appropriate range. 

\begin{figure}
\centering
\includegraphics[width=0.95\textwidth]{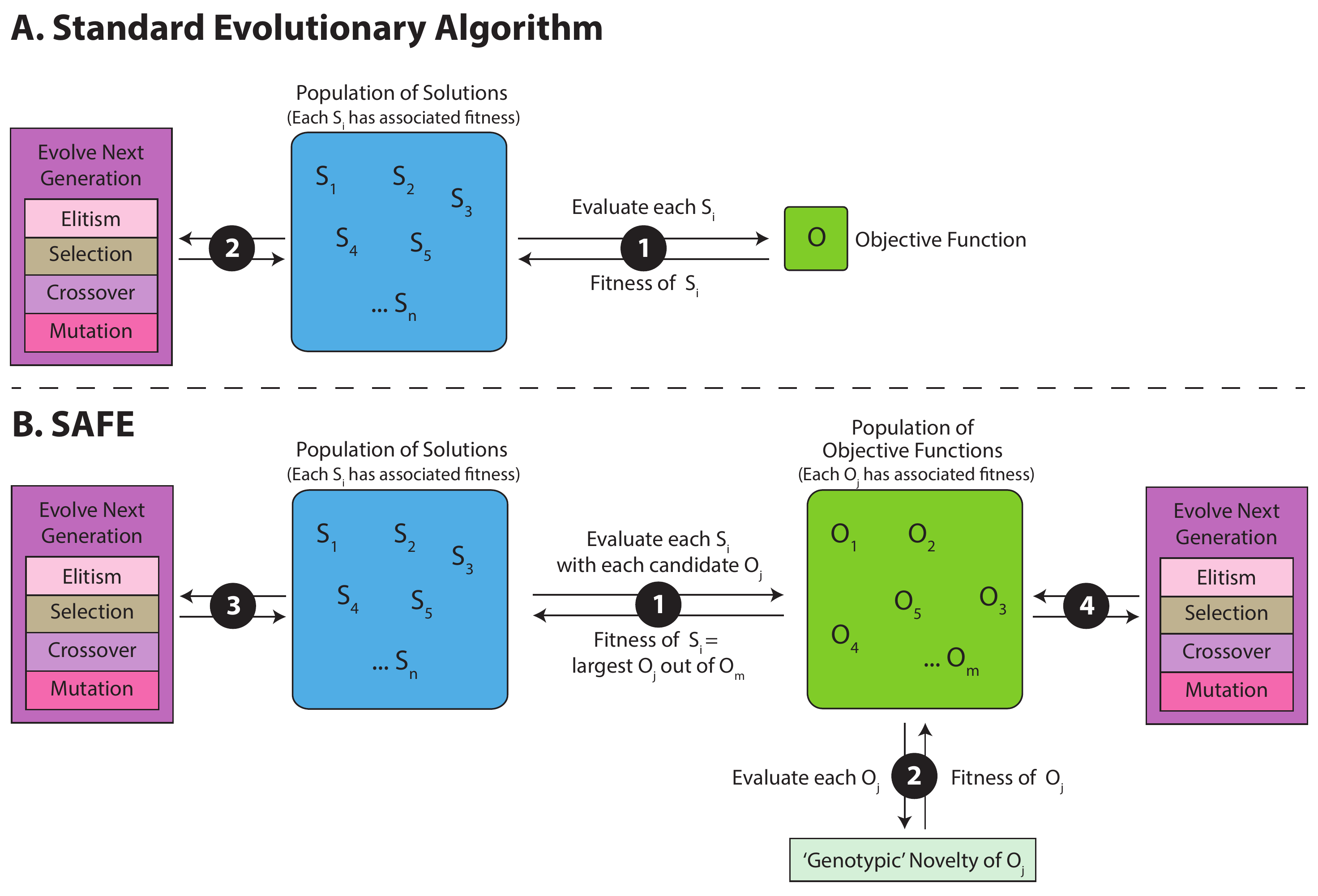}  
\caption{A single generation of SAFE vs. a single generation of a standard evolutionary algorithm. The numbered circles identify sequential steps in the respective algorithms. 
The objective function can comprise a single or multiple objectives. } 
\label{fig:safe}
\end{figure}

\textbf{Fitness.}
Fitness computation is where SAFE dynamics come into play. 
In SAFE, each solution individual, $S_i$, $i \in \{1,\ldots,n\}$ is scored by every candidate objective-function individual, $O_j$, in the current population, $j \in \{1,\ldots,m\}$ (Figure~\ref{fig:safe}B). This is in contrast with a standard evolutionary algorithm, where a population of solutions is evolved but the objective function is fixed (Figure~\ref{fig:safe}A).

A candidate SAFE objective-function individual describes a candidate set of weights, balancing our two objectives of model accuracy and model complexity. The best (highest) of all calculated objective-function scores (out of the set of all individual objective functions in the objective-function population) is then assigned to the individual solution as its fitness value. One can think of other ways to perform this assignment, using average, minimal, maximal, etc' objective-function scores.

As noted above, an objective-function individual is a pair $[a,b]$; specifically, $a$ weights the accuracy component and $b$ weights the complexity component. $O_j(S_i)$ is the fitness score that objective function $O_j$ assigns to solution $S_i$:

\begin{equation}\label{eq:obj}
O_j(S_i) = a_j \times \mathit{accuracy_i} +
           b_j \times  \frac{1}{1 + \mathit{complexity_i}} \, ,
\end{equation}
where 
$\mathit{complexity_i}$ is the number of nodes in the GP tree and
$\mathit{accuracy_i}$ is balanced accuracy as described in \cite{urbanowicz2015exstracs}. This latter measures true/false positives/negatives and combines them by taking into account class imbalances; the value is between $0$ and $1$, with $1$ being perfect accuracy. 

The SAFE code is available at \url{https://github.com/EpistasisLab/}.

\section{Experimental Setup and Results} 
\label{sec:results}

As noted in Section~\ref{sec:intro} our main goal herein is to show that we can forgo the need to a-priori 
perform a balancing act between multiple objectives.
Our experiment thus consists of two extensive sets of runs: a standard EA with fixed weights $a = b = 0.5$, and SAFE, which aims to automatically tune these weights with no need to pre-specify them.

We used GAMETES to generate a variegated suite of datasets, with a mixture of attributes, heritabilities, and problems. For each set of particular settings we generated 30 individual datasets, leading to 30 replicate runs for both SAFE and the standard EA.
The total number of complete evolutionary runs was 840.

We used 5-fold cross-validation: Every simulation run begins by dividing the particular dataset used in that run into 5 random folds. 
Then, the evolutionary algorithm is run 5 times, each time with a training set of 4 folds, leaving aside a hold-out fold---the test set. The best model evolved using the training set is tested with the test set, and its test score recorded. Table~\ref{tab:results} presents our results.

\begin{table}
\centering
\caption{Results:
Each line summarizes 30 replicate runs.
\textbf{Attr}: number of attributes (2 or 3).
\textbf{Herit}: heritability, $\in [0,1]$.
\textbf{Problem}: EDM-1, EDM-2, XOR.
\textbf{Algorithm}: SAFE, standard EA with fixed weights (FIXED). 
Reported algorithm statistics pertain to the 5 best models of each run, evolved for each of the 5 folds---and tested over the (left-out) respective test set.
For every replicate run, compute average of the accuracy and size of the 5 best evolved fold models, and count how many use \textit{all} predictive features (hence a value between 0--5); report average (sd) of all runs: \textbf{Acc}, \textbf{Size}, 
\textbf{Pred}.
}
\resizebox{0.7\textwidth}{!}{
{\begin{tabular}{@{}ccccccc@{}}
\multicolumn{4}{c}{Experimental Settings} & \multicolumn{3}{c}{Evolved Results} \\ \hline
Attr & Herit & Problem & Algorithm & Acc & Size & Pred \\ [3.5pt]\hline\\[-6pt]
2 & 0.1 &  EDM-1 & SAFE & 0.53 (0.03) & 7.69 (1.4) & 3.37 (1.87) \\[1.9pt]
  & & & FIXED & 0.54 (0.02) & 7.69 (2.17) & 4.03 (1.5) \\[1.9pt]
  & & EDM-2 & SAFE & 0.7 (0.02) & 11.45 (5.28) & 5 (0) \\[1.9pt] 
  & & & FIXED & 0.7 (0.03) & 10.35 (5.43) & 5 (0) \\[1.9pt] 
  & 0.2 &  EDM-1 & SAFE & 0.63 (0.03) & 8.01 (2.13) & 4.97 (0.18) \\[1.9pt]
  & & & FIXED & 0.64 (0.02) & 8.16 (3.51) & 4.97 (0.18) \\[1.9pt]
  & & EDM-2 & SAFE & 0.73 (0.03) & 14.52 (9.4) & 5 (0) \\[1.9pt]  
  &  & & FIXED & 0.72 (0.03) & 14.65 (9.3) & 5 (0) \\[1.9pt]
  & 0.3 & EDM-1 & SAFE & 0.68 (0.03)  & 11.59 (6.42) & 5 (0) \\[1.9pt]
  & & & FIXED & 0.68 (0.03) & 10.52 (5.7) & 5 (0) \\[1.9pt]
  & & EDM-2 & SAFE & 0.74 (0.03) & 16.6 (9.99) & 5 (0) \\[1.9pt]  
  & & & FIXED & 0.74 (0.03) & 16.6 (10.82) & 5 (0) \\[1.9pt]
  & 0.4 &  EDM-1 & SAFE & 0.74 (0.02) & 17.33 (8.82) & 5 (0) \\[1.9pt]
  & & & FIXED & 0.73 (0.02) & 22.19 (9.97) & 5 (0) \\[1.9pt]
  & & EDM-2 & SAFE & 0.76 (0.02) & 19.96 (9.45) & 5 (0) \\[1.9pt]  
  & & & FIXED & 0.77 (0.02) & 21.92 (10.21) & 5 (0) \\[1.9pt]
  & 1 & XOR & SAFE & 0.96 (0.02) & 57.8 (15.98) & 5 (0) \\[1.9pt]  
  & & & FIXED & 0.97 (0.02) & 62.43 (17.27) & 5 (0) \\[1.9pt]

3 & 0.1 &  EDM-1 & SAFE & 0.49 (0.02) & 7.85 (1.74) & 0.03 (0.18) \\[1.9pt]
 & & & FIXED & 0.5 (0.02) & 8.77 (2.62) & 0 (0) \\[1.9pt]
  & & EDM-2 & SAFE & 0.5 (0.02) & 8.48 (2.09) & 0.53 (1.31) \\[1.9pt]  
  & & & FIXED & 0.5 (0.02) & 8.04 (1.44) & 0.43 (1.19) \\[1.9pt]
  & 0.2 &  EDM-1 & SAFE & 0.51 (0.02) & 7.93 (1.7) & 0.6 (1.43) \\[1.9pt]
  & & & FIXED & 0.5 (0.02) & 8.31 (1.27) & 0.57 (1.43) \\[1.9pt]
  & & EDM-2 & SAFE & 0.51 (0.02) & 8.53 (2.65) & 1.1 (1.86) \\[1.9pt]  
  & & & FIXED & 0.5 (0.02) & 8.05 (2.16) & 0.93 (1.68) \\[1.9pt]
  & 1 & XOR & SAFE & 0.72 (0.06) & 24.04 (16.48) & 5 (0) \\[1.9pt]  
  & & & FIXED & 0.7 (0.06) & 20.96 (14.14) & 5 (0) \\[1.9pt]


\end{tabular}}\label{tab:results}
}
\end{table}

SAFE's average of weights $a$ and $b$ over all runs was
$a=0.57$ ($\mathit{SD}=0.11$), $b=0.43$ ($\mathit{SD}=0.11$). 

The representation used for both solutions and objective functions is (as with any evolutionary algorithm) of crucial import. While the objective-function representation comprises  2 real values, the solution representation comprises a large function and terminal set. The terminal set is driven by the particular dataset at hand, but one can ask whether a different function set would produce different results. Indeed, we added the functions in Table~\ref{tab:gp2} to the original set of Table~\ref{tab:gp}, and both SAFE and the standard EA were able to improve the results for the 3-way XOR problem (last 2 lines of Table~\ref{tab:results}) to the near-perfect levels of 2-way XOR. This is likely due to the 3-way functions added.

\begin{table}
\centering
\caption{Additions to the function set of Table~\ref{tab:gp}.}
\resizebox{0.9\textwidth}{!}{
{\begin{tabular}{@{}r@{\hskip 10px}l@{\hskip 30px}r@{\hskip 10px}l@{}}
Function & Returns & Function & Returns \\ \hline
\texttt{f\_add3(x,y,z)}  & $x+y+z$ & \texttt{f\_tan10(x)}  & $\mathit{tan}(10*x)$  \\
\texttt{f\_mul3(x,y)}  & $x*y*z$ & \texttt{f\_min3(x,y,z)}  & $\mathit{min}(x,y,z)$ \\
\texttt{f\_mean3(x,y)} & $(x+y+z)/3$ & \texttt{f\_max3(x,y,z)}  & $\mathit{max}(x,y,z)$ \\
\texttt{f\_mul10(x)}  & $x*10$ & \texttt{f\_iflt(x,y,z)}  & \textit{$y$ if $x < 0$ else z} \\
\texttt{f\_div10(x)}  & $x/10$ & \texttt{f\_iflte(x,y,z)}  & \textit{$y$ if $x \leq 0$ else z} \\
\texttt{f\_mod(x,y)}  & \textit{$x\%y$ if $y \neq 0$ else 0} & \texttt{f\_ifgt(x,y,z)}  & \textit{$y$ if $x > 0$ else z}\\
\texttt{f\_abs(x)}  & $\mathit{abs}(x)$ & \texttt{f\_ifgte(x,y,z)}  & \textit{$y$ if $x \geq 0$ else z} \\
\texttt{f\_log(x)}  & \textit{$\mathit{log}(x)$ if $x>0$ else 0} & \texttt{f\_not(x)}  & \textit{1 if $x < 0$ else -1} \\
\texttt{f\_sin(x)}  & $\mathit{sin}(x)$ & \texttt{f\_or(x,y)}  & \textit{1 if $x>0$ or $y>0$ else -1} \\
\texttt{f\_cos(x)}  & $\mathit{cos}(x)$ & \texttt{f\_and(x,y)}  & \textit{1 if $x>0$ and $y>0$ else -1} \\
\texttt{f\_tan(x)}  & $\mathit{tan}(x)$ & \texttt{f\_xor(x,y)}  & \textit{1 if ($x>0$ and $y \leq 0$) or} \\
\texttt{f\_sin10(x)}  & $\mathit{sin}(10*x)$ & & \textit{($x \leq 0$ and $y > 0$) else -1} \\
\texttt{f\_cos10(x)}  & $\mathit{cos}(10*x)$ \\
\end{tabular}}\label{tab:gp2}
}
\verb+ +\vspace{-15pt}\verb+ +
\end{table}

\section{Concluding Remarks}
\label{sec:conc}

We set out to show that when engaged in model discovery we can balance multiple objectives in an automatic manner. Our results show that SAFE was able to do so with no loss in performance---despite being challenged with a more complicated task.

As noted at the outset, this research is still a work-in-progress, as suggested by our focus on complex, but smaller-scaled data analyses . While results are encouraging, there are several paths we envisage for future exploration. 

First, when more objectives are added into the throes, can SAFE handle this? We are currently experimenting with a version of SAFE wherein the objective function is not a linear combination of two objectives, but a full-fledged GP tree combining several objectives. 

Can SAFE not only match but indeed surpass performance of the fixed-weight EA as we expand the data scale and range of model dimensionality and types of pattern complexity?

What other problems in biocomputing can SAFE be applied to? Are there problems and datasets with fuzzy objectives wherein SAFE can help navigate the unknown terrain?

Our final experiment with the addition to the original function set exemplified the well-known importance of representations in EAs. Representation discovery has recently been automated using a coevolutionary algorithm---OMNIREP \cite{sipper2019omnirep,sipper2020art}. 
Can SAFE and OMNIREP be combined beneficially? 

Ultimately, the goal is to facilitate the modeler's job by transferring more tasks to the machine.

\bibliographystyle{splncs04}
\bibliography{bibfile}

\begin{thebibliography}{10}
\providecommand{\url}[1]{\texttt{#1}}
\providecommand{\urlprefix}{URL }
\providecommand{\doi}[1]{https://doi.org/#1}

\bibitem{Bezerra2018}
Bezerra, L.C.T., López-Ibáñez, M., Stützle, T.: A large-scale experimental
  evaluation of high-performing multi- and many-objective evolutionary
  algorithms. Evolutionary Computation  \textbf{26}(4),  621--656 (2018).
  \doi{10.1162/evco\_a\_00217}

\bibitem{cheng2017novel}
Cheng, T., Chen, M., Fleming, P.J., Yang, Z., Gan, S.: A novel hybrid teaching
  learning based multi-objective particle swarm optimization. Neurocomputing
  \textbf{222},  11--25 (2017)

\bibitem{Darwin:1859}
Darwin, C.R.: On the Origin of Species by Means of Natural Selection, or the
  Preservation of Favoured Races in the Struggle for Life. John Murray, London
  (1859)

\bibitem{deb2002fast}
Deb, K., Pratap, A., Agarwal, S., Meyarivan, T.: A fast and elitist
  multiobjective genetic algorithm: {NSGA-II}. IEEE Transactions on
  Evolutionary Computation  \textbf{6}(2),  182--197 (2002)

\bibitem{han2018improved}
Han, F., Sun, Y.W.T., Ling, Q.H.: An improved multiobjective quantum-behaved
  particle swarm optimization based on double search strategy and circular
  transposon mechanism. Complexity  \textbf{2018} (2018)

\bibitem{huband2006review}
Huband, S., Hingston, P., Barone, L., While, L.: A review of multiobjective
  test problems and a scalable test problem toolkit. IEEE Transactions on
  Evolutionary Computation  \textbf{10}(5),  477--506 (2006)

\bibitem{koza92}
Koza, J.R.: Genetic Programming: On the Programming of Computers by Means of
  Natural Selection. MIT Press, Cambridge, MA, USA (1992)

\bibitem{metevier2019lexicase}
Metevier, B., Saini, A.K., Spector, L.: Lexicase selection beyond genetic
  programming. In: Banzhaf, W., Spector, L., Sheneman, L. (eds.) Genetic
  Programming Theory and Practice XVI, pp. 123--136. Springer (2019)

\bibitem{ni2012use}
Ni, J., Drieberg, R.H., Rockett, P.I.: The use of an analytic quotient operator
  in genetic programming. IEEE Transactions on Evolutionary Computation
  \textbf{17}(1),  146--152 (2012)

\bibitem{Pena:2001}
Pena-Reyes, C.A., Sipper, M.: Fuzzy {CoCo}: A cooperative-coevolutionary
  approach to fuzzy modeling. IEEE Transactions on Fuzzy Systems
  \textbf{9}(5),  727--737 (2001)

\bibitem{Sipper2018par}
Sipper, M., Fu, W., Ahuja, K., Moore, J.H.: Investigating the parameter space
  of evolutionary algorithms. BioData Mining  \textbf{11}(2),  1--14 (2018)

\bibitem{sipper2019omnirep}
Sipper, M., Moore, J.H.: {OMNIREP}: Originating meaning by coevolving encodings
  and representations. Memetic Computing pp. 1--11 (2019)

\bibitem{Sipper2019multiobj}
Sipper, M., Moore, J.H., Urbanowicz, R.J.: Solution and fitness evolution
  ({SAFE}): A study of multiobjective problems. In: Proceedings of 2019 IEEE
  Congress on Evolutionary Computation. pp. 1868--1874. IEEE (2019)

\bibitem{Sipper2019robot}
Sipper, M., Moore, J.H., Urbanowicz, R.J.: Solution and fitness evolution
  ({SAFE}): Coevolving solutions and their objective functions. In: Sekanina,
  L., Hu, T., Louren{\c{c}}o, N., Richter, H., Garc{\'i}a-S{\'a}nchez, P.
  (eds.) Genetic Programming. pp. 146--161. Springer International Publishing,
  Cham (2019)

\bibitem{sipper2020art}
Sipper, M., Moore, J.H., Urbanowicz, R.J.: Coevolving artistic images using
  {OMNIREP}. In: Romero, J., Ek{\'a}rt, A., Martins, T., Correia, J. (eds.)
  Artificial Intelligence in Music, Sound, Art and Design. pp. 165--178.
  Springer International Publishing, Cham (2020)

\bibitem{spector2012assessment}
Spector, L.: Assessment of problem modality by differential performance of
  lexicase selection in genetic programming: A preliminary report. In:
  Proceedings of the 14th Annual Conference Companion on Genetic and
  Evolutionary Computation. pp. 401--408. ACM (2012)

\bibitem{urbanowicz2012predicting}
Urbanowicz, R.J., Kiralis, J., Fisher, J.M., Moore, J.H.: Predicting the
  difficulty of pure, strict, epistatic models: metrics for simulated model
  selection. BioData mining  \textbf{5}(1), ~15 (2012)

\bibitem{urbanowicz2012gametes}
Urbanowicz, R.J., Kiralis, J., Sinnott-Armstrong, N.A., Heberling, T., Fisher,
  J.M., Moore, J.H.: {GAMETES}: a fast, direct algorithm for generating pure,
  strict, epistatic models with random architectures. BioData mining
  \textbf{5}(1), ~16 (2012)

\bibitem{urbanowicz2015exstracs}
Urbanowicz, R.J., Moore, J.H.: {ExSTraCS} 2.0: Description and evaluation of a
  scalable learning classifier system. Evolutionary Intelligence
  \textbf{8}(2-3),  89--116 (2015)

\bibitem{zhou2011multiobjective}
Zhou, A., Qu, B.Y., Li, H., Zhao, S.Z., Suganthan, P.N., Zhang, Q.:
  Multiobjective evolutionary algorithms: A survey of the state of the art.
  Swarm and Evolutionary Computation  \textbf{1}(1),  32--49 (2011)

\bibitem{zitzler2000comparison}
Zitzler, E., Deb, K., Thiele, L.: Comparison of multiobjective evolutionary
  algorithms: Empirical results. Evolutionary computation  \textbf{8}(2),
  173--195 (2000)

\end{thebibliography}

\end{document}